\documentclass{article}

\usepackage[preprint]{neurips_2019}

\usepackage[utf8]{inputenc} 
\usepackage[T1]{fontenc}    
\usepackage{hyperref}       
\usepackage{url}            
\usepackage{booktabs}       
\usepackage{amsfonts}       
\usepackage{nicefrac}       
\usepackage{microtype}      
\usepackage{xcolor}
\usepackage[pdftex]{graphicx}
\usepackage{caption}
\usepackage{subcaption}
\usepackage{subcaption}
\usepackage{amsmath}
\usepackage{soul}
\usepackage{booktabs}
\usepackage{float}
\usepackage{graphicx}
\usepackage[compact]{titlesec}
\usepackage{multicol}
\titlespacing{\section}{0pt}{*0}{*0}
\titlespacing{\subsection}{0pt}{*0}{*0}
\titlespacing{\subsubsection}{0pt}{*0}{*0}
\graphicspath{ {.} }
\title{Fully Bayesian Recurrent Neural Networks for Safe Reinforcement Learning}

\author{
  Matt ~Benatan\\
  IBM Research UK \\
  Sci-Tech Daresbury\\
  Warrington, UK. \\
  \texttt{matthew.benatan@ibm.com} \\
  \And
  Edward O.~Pyzer-Knapp \\
  IBM Research UK \\
  Sci-Tech Daresbury\\
  Warrington, UK. \\
  \texttt{epyzerk3@uk.ibm.com} \\
}
\begin{document}

\maketitle


\begin{abstract}
Reinforcement Learning (RL) has demonstrated state-of-the-art results in a number of autonomous system applications, however many of the underlying algorithms rely on black-box predictions. This results in poor explainability of the behaviour of these systems, raising concerns as to their use in safety-critical applications. Recent work has demonstrated that uncertainty-aware models exhibit more cautious behaviours through the incorporation of model uncertainty estimates. In this work, we build on Probabilistic Backpropagation to introduce a fully Bayesian Recurrent Neural Network architecture. We apply this within a Safe RL scenario, and demonstrate that the proposed method significantly outperforms a popular approach for obtaining model uncertainties in collision avoidance tasks. Furthermore, we demonstrate that the proposed approach requires less training and is far more efficient than the current leading method, both in terms of compute resource and memory footprint.

\end{abstract}

\section{Introduction}
Reinforcement Learning (RL) has achieved state-of-the-art results in a variety of applications, from Atari games \citep{mnih2013playing} to autonomous vehicles \citep{zhang2016learning, shalev2016safe}. The models underlying these systems are often black-box in nature, and do not provide estimates of model uncertainty. This results in over-confident predictions on out-of-distribution data, resulting in poor performance in novel data scenarios \citep{amodei2016concrete, lakshminarayanan2017simple, hendrycks2016baseline}. Recent work \citep{kahn2017uncertainty, lutjens2018safe} demonstrates that more cautious behaviours can be attained by incorporating model uncertainty estimates into the decision making processes of RL agents. In their work, Lütjens \textit{et al.} demonstrate that a Long-Short-Term-Memory (LSTM) ensemble using MC Dropout \citep{gal_dropout_2015} is able to approximate model uncertainties. While this demonstrates a clear advantage over a standard LSTM with no uncertainty estimates, the accuracy of their approach's uncertainty estimates is tied to the size of the ensemble and the number of dropout forward passes executed. Thus, obtaining high quality model uncertainty estimates can quickly become demanding for both compute and memory resources.

An alternative to the approximate Bayesian inference provided by MC Dropout and ensemble methods is Probabilistic Backpropagation (PBP)  \citep{hernandez-lobato_probabilistic_2015} - a fully Bayesian method for training Bayesian Neural Networks (BNNs), which thus produces fully-Bayesian model uncertainty estimates. Unlike ensemble-based methods \citep{lakshminarayanan2017simple}, PBP provides a probabilistic neural network architecture at only twice the memory footprint of its non-probabilistic equivalent. As PBP is fully Bayesian, it only requires training a single network, and inference only requires one forward pass - making it far less computationally intensive than ensemble \citep{lakshminarayanan2017simple}, MC Dropout \citep{gal_dropout_2015}, or combined \citep{lutjens2018safe} methods.

In this paper we leverage PBP to produce a probabilistic variant of a standard Recurrent Neural Network (RNN). We apply this to a collision avoidance task, and demonstrate that the PBP-RNN exhibits competitive performance when compared with the MC Dropout Ensemble (MDE) described in \citep{lutjens2018safe}, and achieves higher-quality model uncertainty estimates.

\section{Related Work}
\subsection{Safe Reinforcement Learning}
Safe RL involves learning policies which maximize performance criteria, e.g. reward, while accounting for safety constraints \citep{garcia2015comprehensive, berkenkamp2017safe}, and is a field of study that is becoming increasingly important as more and more automated systems are being designed to operate in safety-critical situations \citep{zhang2016learning, chen2019model, shalev2016safe}. A key issue with many existing approaches is their lack of uncertainty quantification - agents' inability to quantify what they `don't know'. By incorporating uncertainty, RL agents can make better decisions by opting for actions for which they are more confident of a positive outcome. Incorporating this principal in safety-constrained tasks results in agents choosing safer actions \citep{kahn2017uncertainty, lutjens2018safe}, and is crucial for making RL feasible for safety-critical scenarios.

Existing work on Safe RL has typically aimed to discover uncertainty in the environment or model \citep{garcia2015comprehensive} - with the former commonly being the goal in the case of risk-sensitive RL (RSRL) \citep{mihatsch2002risk, shen2014risk}. In this work we focus on model uncertainty, with the aim of quantifying the model's confidence on out-of-distribution data and using this uncertainty quantification to produce safer decisions.

\subsection{Probabilistic Neural Networks}
There are a variety of approaches for modeling distributions and producing accurate uncertainty estimates \citep{williams2006gaussian, ghahramani2015probabilistic}, however many of these methods do not scale well to large datasets. Over recent years neural network-based methods have demonstrated state-of-the-art performance on a wide range of tasks \citep{cho2014learning, he2016deep, ledig2017photo}, partly due to their ability to leverage large amounts of data. This has helped to drive interest in the development of a variety of probabilistic neural network approaches, which are capable of modeling uncertainty while also scaling to large datasets \citep{gal_dropout_2015, hernandez-lobato_probabilistic_2015, snoek_scalable_2015}.

In probabilistic neural networks, network weights are modeled as distributions, rather than as point estimates, allowing the network to encode model uncertainty. If we consider $\textbf{y}$ to be an $N$-dimensional vector of targets $y_n$, and $\textbf{X}$ to be an $N \times D$ matrix of features $\textbf{x}_n$, then we can define the likelihood for $\textbf{y}$ given the network weights $W$, $\textbf{X}$, and noise precision $\gamma$ as:

\begin{equation}
    p(\textbf{y}|W, \textbf{X}, \gamma) = \prod_{n = 1}^N \mathcal{N} (y_n | f(\textbf{x}_n;W), \gamma^{-1})
\end{equation}

 We specify a Gaussian prior for each weight in our network, giving us:

\begin{equation}
   p(W|\lambda_{p}) = \prod_{l=1}^L \prod_{i=1}^{V_{l}} \prod_{j=1}^{V_{l-1}+1} \mathcal{N}(w_{ij, l}|0, \lambda_{p}^{-1})
\end{equation}

where $w_{ij,l}$ denotes a weight in weight matrix $\textbf{W}_l$ at layer $l$, $V_l$ is the number of neurons in layer $l$, and $\lambda_{p}$ is a precision parameter. For additional details pertaining to $\lambda_{p}$ and its hyper-prior, please see \citep{hernandez-lobato_probabilistic_2015}. Using the above definition, we can obtain the posterior distribution for parameters $W$, $\gamma$ and $\lambda_{p}$ by applying Bayes' rule:

\begin{equation}
    p(W, \gamma, \lambda_{p} | \mathcal{D}) = \frac{p(\textbf{y}|W, \textbf{X}, \lambda_{p})p(W|\lambda_{p})p(\lambda_{p})p(\gamma)}{p(\textbf{y}|\textbf{X})}
\end{equation}

where $\mathcal{D} = (\textbf{X}, \textbf{y})$. Predictions for some output $y_\star$ can then be obtained through:

\begin{equation}
    p(y_\star | \textbf{x}_\star, \mathcal{D}) = \int  p(y_\star | \textbf{x}_\star, W, \gamma) p(W, \gamma, \lambda_{p} | \mathcal{D}) d\gamma d\lambda_{p} dW
\end{equation}

As this integral is computationally intractable, several approximations have been proposed \citep{hernandez-lobato_probabilistic_2015, gal_dropout_2015, ghosh_assumed_2016}.  In this work, we use PBP to train a fully Bayesian neural network, as opposed to the approximation provided by MC Dropout or ensemble-based methods.

\subsection{Probabilistic Backpropagation}
The PBP network can be viewed as a modification of a standard Multilayer Perceptron (MLP) wherein each weight $w_{ij,l} \in \textbf{W}_l$ is defined by a one dimensional Gaussian and correspondingly is represented by two weights - a mean $m_{ij,l}$ and a variance $v_{ij,l}$.

Similarly to standard backpropagation, PBP training consists of two phases. The first phase comprises forward propagation of the input features through the network to obtain the marginal log-likelihood, $logZ$ (instead of loss, which is used in typical backpropagation). The gradients of $logZ$ with respect to the mean and variance weights are then backpropagated using reverse-mode differentiation, and the resulting derivatives are used to update the mean and variance weights of the network.

The update rule for PBP obtains the parameters of the new Gaussian beliefs $q^{new}(w) = \mathcal{N}(w|m^{new}, v^{new})$ that minimize the Kullback-Leibler (KL) divergence between our beliefs $s$ and $q^{new}$ as a function of $m$, $v$ and the gradient of $logZ$:

\begin{equation}
    m^{new} = m + v\frac{\partial logZ}{\partial m}
\end{equation}

\begin{equation}
    v^{new} = v - v^2 \Bigg[ \frac{\partial logZ}{\partial m}^2 - 2 \frac{\partial logZ}{\partial v} \Bigg]
\end{equation}

These rules ensure moment matching between $q^{new}$ and s, guaranteeing that the distributions have the same mean and variance. These are the update equations used in the PBP-RNN described in this paper. For further details on PBP, please see \citep{hernandez-lobato_probabilistic_2015}.

\section{Probabilistic Backpropagation for Recurrent Neural Networks}
Recurrent Neural Networks (RNNs), and specifically LSTMs, are the current state-of-the-art for dynamic obstacle avoidance tasks \citep{alahi2016social, lutjens2018safe, vemula2018social}. This is due to their ability to encode contextual dependencies, allowing them to accurately model the temporal patterns of dynamic obstacles. As such, we have chosen to use an RNN for our collision prediction network. In order to provide fully-Bayesian model uncertainty estimates, we adapt a standard RNN architecture to use Probabilistic Backpropagation (PBP).

A standard RNN consists of two weight matrices per layer, the weight matrix $\textbf{W}_x$, which is equivalent to the weight matrices used in a standard MLP, and the weight matrix $\textbf{W}_h$, which holds the transition weights. For a given input $\textbf{x}_t$ at a time step $t$, the output $\textbf{y}_t$ and hidden state $\textbf{h}_t$ at step $t$ are computed as:

\begin{equation}
    \textbf{h}_t = \textbf{y}_t = f(\textbf{W}_h h_{t-1} + \textbf{W}_x\textbf{x}_t)
\end{equation}

where $f()$ is some activation function and $\textbf{h}_{t-1}$ is the previous hidden state.

For the PBP-RNN, we apply the PBP treatment of a standard MLP to an RNN: each of the weights $w_h \in \textbf{W}_h$ and $w_x \in \textbf{W}_x$ are represented by mean and variance weights, resulting in four weight matrices in place of the two standard RNN matrices: $\textbf{W}_{m}$, $\textbf{W}_{v}$, $\textbf{W}_{hm}$ and $\textbf{W}_{hv}$. This produces two outputs from the network output layer $L$ at each time step $t$ - a mean $m_{L,t}$ and a variance $v_{L,t}$:

\begin{equation}
    m_{L, t} = \textbf{W}_{hm} \textbf{h}_{mt-1} + \textbf{W}_{m}\textbf{x}_t
\end{equation}

\begin{equation}
    v_{L, t} = \textbf{W}_{hv} \textbf{h}_{vt-1} + \textbf{W}_{v}\textbf{v}_t
\end{equation}



Updates to the network at step $t$ are computed using Truncated Backpropagation Through Time (TBTT) \citep{boden2002guide} in combination with the standard PBP update procedure, whereby the network is `unrolled' by some $T$ time steps, and the weights are updated in reverse order - from time step $T$ back to $t=1$. First, the marginal log likelihoods for each time step are computed:

\begin{equation}
    logZ_t = -0.5 \frac{\log{v_t} + (y_t - m_{l,t})^2}{v_t}
\end{equation}

These are then used to update the weights in each of the four weight matrices per layer, as per the typical PBP update rule:

\begin{align}
    m_t^{new} &= m_t + v_t\frac{\partial logZ_t}{\partial m_t} \; \forall \; m_t \; \in \; \textbf{W}_{m}\\
    m_{ht}^{new} &= m_{ht} + v_{ht}\frac{\partial logZ_t}{\partial m_{ht}} \; \forall \; m_{ht} \; \in \; \textbf{W}_{hm}\\
    v_t^{new} &= v_t - v_t^2 \Bigg[ \frac{\partial logZ_t}{\partial m_t}^2 - 2 \frac{\partial logZ_t}{\partial v_t} \Bigg] \; \forall \; v_t \; \in \; \textbf{W}_{v}\\
    v_{ht}^{new} &= v_{ht} - v_{ht}^2 \Bigg[ \frac{\partial logZ_t}{\partial m_{ht}}^2 - 2 \frac{\partial logZ_t}{\partial v_{ht}} \Bigg] \; \forall \; v_{ht} \; \in \; \textbf{W}_{hv}
\end{align}

For incorporating prior factors into $q$, we follow the same procedure for updating the $\alpha$ and $\beta$ parameters described in \citep{hernandez-lobato_probabilistic_2015}, but substitute the likelihood $Z$ for the mean of $Z$ over all time steps:

\begin{equation}
    Z = \frac{{\sum^T_{t=1} Z_t}}{T}
\end{equation}

The same procedure is followed for parameters $Z_1$ and $Z_2$ used in the $\alpha$ and $\beta$ updates for standard PBP, as described in \citep{hernandez-lobato_probabilistic_2015}. As with standard BPTT, this process is repeated for each time step $t \in T$ for each mini-batch within each epoch.


\section{Methods}
Many tasks used for evaluating the performance of RL methods, such as arcade learning environments, would not obviously benefit from a safe RL approach. In order to transparently and reproducibly assess the effects of our approach to uncertainty quantification, we decided that the tasks which would be used for this analysis would include the following qualities:
\begin{itemize}
    \item The task should reflect a real-world application of RL, rather than a trivial toy problem.
    \item The task should be simple enough that the results can be clearly interpreted.
    \item The task should facilitate a comprehensive combination of test cases for evaluating the impact of using uncertainty quantification, including exposing the agent to novel behaviours, noise, and missing data.
\end{itemize}
We felt that the task of collision avoidance, as in \citep{lutjens2018safe}, fulfilled this criteria, and was a sensible choice to comprehensively evaluate our methods. As such, in this work we use the PBP-RNN architecture described above for a collision prediction network applied to a collision avoidance task. This network is used within a Model Predictive Controller (MPC) which selects a motion primitive $u$ from a set of motion primitives $U$ as in \citep{lutjens2018safe}. Similarly to \citep{lutjens2018safe}, $U$ contains 11 discrete motion primitives spanning heading angles $\alpha_h \in [-\frac{\pi}{5},\frac{\pi}{5}]$ of length $h=0.05$, and the MPC chooses the lowest-cost motion primitive according to the following criteria:

\begin{equation}
    u^*_{t:t+h} = argmin(\lambda_vV^i_{coll} + \lambda_cP^i_{coll} + \lambda_dd_{goal})
\end{equation}

where $P^i_{coll}$ is the collision prediction for the motion primitive at $i$, $V^i_{coll}$ denotes the variance associated with the collision prediction, $d_{goal}$ is the distance to the goal, and $\lambda_v$, $\lambda_c$, $\lambda_d$ are their respective coefficients. In this work, we use an epsilon greedy policy \citep{mnih2015human}, decreasing $\epsilon$ by $\frac{\epsilon}{50}$ after each episode, to continue adding new experience while monotonically decreasing the probability of random actions. Previous work demonstrates that starting with low $\lambda_v$ and increasing $\lambda_v$ over time is helpful for escaping local minima \citep{lutjens2018safe}. As such, we multiply $\lambda_v$ by $1-\epsilon$, so that $\lambda_v$ increases as $\epsilon$ decreases.


\begin{figure}[H]
    \centering
    \includegraphics[width=0.5\textwidth]{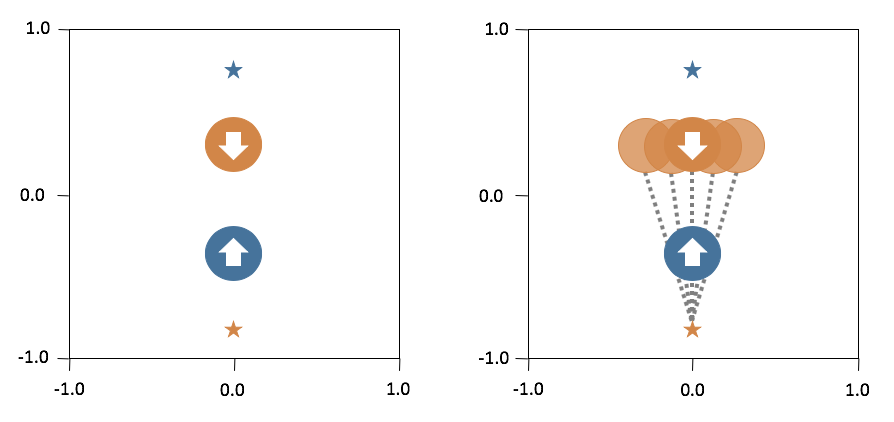}
    \caption{Diagram of initial environment settings for training distribution (left) and test distribution (right). Top (orange) circle: dynamic obstacle. Bottom (blue) circle: agent. Bottom star: goal for dynamic obstacle. Top star: goal for agent.}
\end{figure}

\subsection{Environment}

For the collision avoidance task, we use a multi-agent OpenAI Gym \citep{brockman2016openai} environment based on \citep{lowe2017multi} with two agents. The agents start each episode facing eachother, and each agent is assigned a goal, as illustrated in Figure 1. One agent acts as a dynamic obstacle, following a collaborative Reciprocal Velocity Object (RVO) policy \citep{van2011reciprocal}, while the other uses the MPC and collision prediction network. The observation $\textbf{o}_t$ at time $t$ is defined as:
\begin{equation}
    \textbf{o}_t = [a_{1p,t}, a_{1v,t}; a_{2p,t}, a_{2v,t}, u_{t}]
\end{equation}

where $[a_{1p}, a_{1v}]$ and $[a_{2p}, a_{2v}]$ are the positions and velocities for agents 1 and 2 respectively and $u_{t}$ is the evaluated motion primitive. For the input to the RNN (and the LSTM, which we compare against), we feed the input \textbf{o}, which comprises $l = 8$ observations, from $t - l$ to $t$. After each episode, all inputs are collated into a set of input features $\textbf{X}$, and a set of labels $\textbf{y}$ is populated with the collision label, $y = [0, 1]$, for the episode. These are collated into an experience pool of examples, from which samples are randomly drawn to train the collision prediction network.


The results reported here were obtained after $\epsilon$ reached $0.1$, at which point it was set to $0$, and the test cases were run with no further training cycles. While $\epsilon > 0$, the dynamic obstacle starts in the same location each episode, $pos_{xy,o} = [0, 0.25]$, and the agent starts at $pos_{xy,a} = [0, -0.25]$. Once $\epsilon$ is set to $0$, we obtain results for 20 episodes using this initialization, after which the starting position of the dynamic obstacle is randomly generated for $pos_{y,o}$ from the range $-0.25$ to $0.25$, as illustrated in Figure 1. At this point we switch to a non-collaborative policy, for which the obstacle no longer tries to avoid the agent, and simply moves in a straight-line towards its goal. This combination of random initialization and policy switching for the dynamic obstacle produces novel dynamic object behaviour which is used for testing.


\subsection{Network Parameters}
Using previous work as a guide \citep{lutjens2018safe}, the LSTM Ensemble consists of 5 single-layer LSTM networks each with 16 units with linear activations and a Rectified Linear Unit (ReLU) activation on the output. This is trained using MSE loss and Adam optimization with an initial learning rate of 0.001. For inference (as in \citep{lutjens2018safe}), we execute 20 forward passes per network with different dropout masks (setting dropout $p = 0.7$) from which the sample mean and variance are drawn from the resulting distribution of 100 predictions. We use an equivalent architecture for the PBP-RNN - a single layer network consisting of 16 units with linear activations. For both networks we set the number of time steps $T$ equal to the sequence length $l_s$. For shorter sequences (for the first 7 time steps in each episode) we zero pad the input up to $T$.



\subsection{Training Procedure}
For both networks evaluated here we used TensorFlow on a single Power 8+ compute node. During the training cycles, the dynamic obstacle follows a collaborative RVO policy. We first run 100 episodes to seed the experience pool, for which the agent selects random actions. Following this we draw 500 random samples from the pool to train the collision prediction network. We then follow a standard observe-act-train procedure, repeating training after every 10 episodes. To balance the training data, we draw half of the samples at random from examples where collision labels are $y=0$ and half from a pool where collision labels are $y=1$.

For PBP, we run an initial phase of 5 epochs of training after the first 100 episodes, and 2 epochs of training after each subsequent set of 10 episodes. This relatively small number of training epochs is guided by empirical evidence and prior work demonstrating that PBP requires relatively few epochs for training \citep{benatan2018practical, hernandez-lobato_probabilistic_2015}. We found that the MDE performed better with comparatively more epochs of training, and so trained this for 100 epochs after the first 100 episodes, followed by 10 epochs of training after each subsequent set of 10 episodes.

For the MPC, we use the same values for the $\lambda$ parameters as recommended in \citep{lutjens2018safe} for both the MDE and PBP-RNN, setting $\lambda_c = 25$, $\lambda_v = 200$ and $\lambda_d = 3$.


\section{Results}

\begin{table}[]
\centering
\caption{Means ($\mu$) and variances ($\sigma^2$) of recorded variances for PBP-RNN and MDE for the training distribution and three out-of-distribution test cases}
\label{tab:my-table}
\begin{tabular}{lccccl}
 & \multicolumn{1}{l}{} & \multicolumn{1}{l}{} & \multicolumn{1}{l}{} & \multicolumn{1}{l}{} &  \\ \cline{1-5}
 & PBP-RNN $\mu$ & MDE$\mu$ & PBP-RNN $\sigma^2$ & MDE $\sigma^2$ &  \\ \cline{1-5}
                        Train & 0.002 & 0.001 & 0.002 & 0.001 &  \\
                        Novel & 0.003 & 0.001 & 0.002 & 0.0002 &  \\
Novel + noise ($\lambda_\xi = 0.005$) & 0.006 & 0.001 & 0.005 & 0.0003 &  \\
Novel + dropped obs. ($n_{dropped} = 5$) & 0.039 & 0.007 & 0.044 & 0.003 &  \\ \cline{1-5}
 &  &  &  &  &
\end{tabular}
\vspace{-4mm}
\end{table}

\begin{table}[]
\centering
\caption{Means ($\mu$) and variances ($\sigma^2$) of recorded log-likelihoods for PBP-RNN and MDE for the training distribution and three out-of-distribution test cases}
\label{tab:my-table}
\begin{tabular}{lccccl}
 & \multicolumn{1}{l}{} & \multicolumn{1}{l}{} & \multicolumn{1}{l}{} & \multicolumn{1}{l}{} &  \\ \cline{1-5}
 & PBP-RNN $\mu$ & MDE$\mu$ & PBP-RNN $\sigma^2$ & MDE $\sigma^2$ &  \\ \cline{1-5}
Train & \textbf{0.685 }& -2.1$\times 10^5$ & \textbf{0.108} & 3.8$\times 10^5$ &  \\
Novel & \textbf{-0.555} & -6.2$\times 10^4$ & \textbf{1.048} & 1.7$\times 10^5$ &  \\
Novel + noise ($\lambda_\xi = 0.005$) & \textbf{-2.479} & -8.6$\times 10^4$ & \textbf{1.440} & 2.5$\times 10^5$ &  \\
Novel + dropped obs. ($n_{dropped} = 5$) & \textbf{-8.760} & -66.871 & \textbf{4.707} & 27.120 &  \\ \cline{1-5}
 &  &  &  &  &
\end{tabular}
\vspace{-4mm}
\end{table}


\begin{figure}[H]
    \centering
    \includegraphics[width=1.0\textwidth]{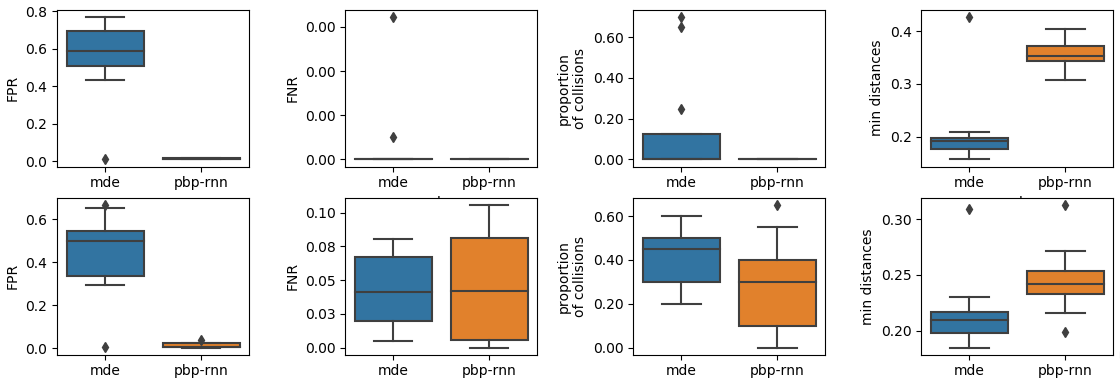}
    \caption{Box-and-whisker plots of results for collision detection network performance. Top: results from training distribution. Bottom: results from test (novel) distribution.}
\end{figure}

\subsection{Collision Detection Network Performance}
Following the completion of all training cycles, we collect a set of in-distribution (training) and a set of out-of-distribution (novel) test data, each for 20 episodes - the entire process is repeated 10 times. For the out-of-distribution data, we randomly initialize the dynamic obstacle as described in Section 4.1. We use this data to evaluate the performance of the collision prediction network through obtaining the log-likelihood, false positive rate (FPR) (the number of times a collision is predicted, but not encountered) and false negative rate (FNR) (the number of times a collision occurred when one was not predicted). Additionally, for all non-collision episodes, we record the minimum distances between the agent and obstacle in order to build an impression of the caution exhibited by the agent.

As demonstrated in Figure 2, the PBP-RNN achieves substantially better results on the training distribution, with 0 false-positives, false-negatives, and collisions. As would be expected, we see these figures degrade somewhat in the case of novel data for both the PBP-RNN and the MDE; however the PBP-RNN continues to demonstrate better overall performance, with a significantly lower FPR and fewer collisions.

The minimum distance plots in Figure 2 (far right top and bottom) demonstrate that the agent using the PBP-RNN leaves a greater margin when passing the obstacle than the agent using the MDE; indicating more cautious behavior. This is supported by the variance results in Table 1, which show that the PBP-RNN's variance values increase between the training and novel scenarios, whereas this isn't the case for MDE. This indicates that the PBP-RNN has a comparatively higher quality model of uncertainty, as its variances increase with the degree of novelty in the observations - thus resulting in more cautious behaviour.

\setlength{\abovedisplayskip}{3pt}
\setlength{\abovedisplayshortskip}{3pt}
\setlength{\belowdisplayskip}{3pt}
\setlength{\belowdisplayshortskip}{3pt}

\subsection{Collision Avoidance with Noise and Dropped Observations}
Uncertainty-based methods have demonstrated advantageous performance in the presence of noise or dropped observations due to their ability to leverage model uncertainty when selecting actions \citep{lutjens2018safe}. Here, we investigate the performance of the MDE and PBP-RNN approaches with varying levels of additive noise by adding a matrix of noise terms $\boldsymbol{\xi}$ multiplied by an incrementally increasing noise coefficient $\lambda_{\xi}$ to our matrix of observations:
\begin{equation}
    \textbf{o}_{t\xi} = \textbf{o}_t + \boldsymbol{\xi} \lambda_{\xi}
\end{equation}
where $\boldsymbol{\xi}$ is a matrix of values each generated from the distribution $\mathcal{N}(0, 1)$. For each method, we run 10 episodes for each value of $\lambda_{\xi}$ after the initial training phase, and repeat this process 10 times. We again use the non-collaborative policy for the dynamic obstacle. As Figure 3 demonstrates, the PBP-RNN exhibits better robustness to noise, with fewer collisions on average. This performance can again be explained by the change in the PBP-RNN's variance output as the level of noise increases. This is demonstrated in the bottom plot of Figure 3, which shows $\sigma^2$ steadily increasing for the PBP-RNN, whereas this is not the case for MDE - again indicating that the PBP-RNN has a better model of uncertainty.

In the final test case, we combine the non-collaborative policy with dropped observations - whereby, for increasing values of $n_{dropped}$, between 1 and 8 observations in the sequence are randomly selected and set to zero, simulating the kind of behaviour that may occur in electronic sensor systems. The PBP-RNN again exhibits better performance when compared with the MDE, as shown in Figure 4, and the same underlying theme is evident: the $\sigma^2$ values for the PBP-RNN increase with the number of dropped observations, whereas these remain very small (although do increase marginally) for the MDE.

Tables 1 and 2 further validate the model quality of the PBP-RNN with respect to the MDE. In Table 1, we see that the PBP-RNN's mean uncertainty increases as the test scenarios deviate from the training distribution. While the MDE's variance does eventually increase, it requires the combination of the non-collaborative policy and a significant number of dropped observations for this to occur. Crucially, Table 2 illustrates that the log-likelihood values for the PBP-RNN and MDE differ significantly, with the PBP-RNN obtaining high and consistent log-likelihood scores, while MDE achieves low and inconsistent log-likelihoods. These poor log-likelihoods are a product of MDE's over-confident predictions due to its lack of a descriptive posterior - a known drawback of MC Dropout \citep{lutjens2018safe, pearce2018bayesian}, which is clearly documented here in the variance and log-likelihood values of our results.

\begin{figure}[H]
    \centering
    \includegraphics[width=0.7\textwidth]{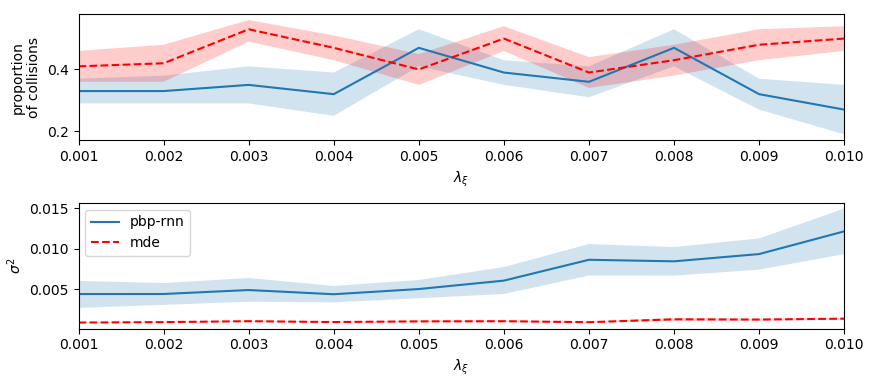}
    \caption{Plots depicting results for collision avoidance in increasingly noisy conditions. Top: proportion of collisions as $\lambda_\xi$ is increased. Bottom: $\sigma^2$ (variance) values as $\lambda_\xi$ is increased. Shaded area denotes 95\% confidence interval.}
\end{figure}

\begin{figure}[H]
    \centering
    \includegraphics[width=0.7\textwidth]{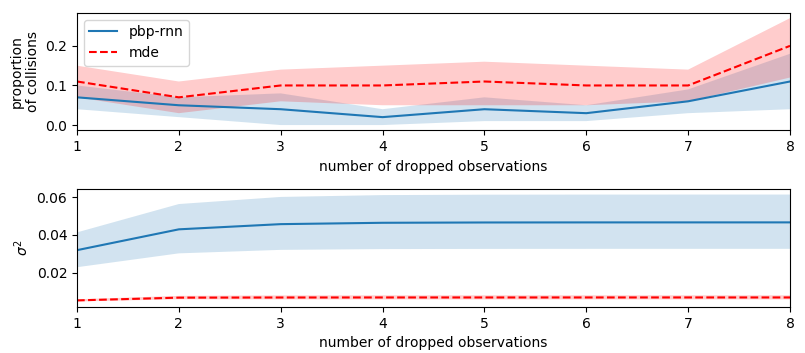}
    \caption{Plots depicting results for collision avoidance with increasing numbers of dropped observations. Top: proportion of collisions as the number of dropped observations is increased. Bottom: $\sigma^2$ (variance) values as the number of dropped observations is increased. Shaded area denotes 95\% confidence interval.}
\end{figure}

\subsection{Computational Considerations}
We analyzed compute time for inference with both the PBP-RNN and the MC Dropout Ensemble. For MDE, we ran the forward passes serially and recorded a mean inference time of $119.8\pm3.2\mathrm{ms}$. In the case of the PBP-RNN, we recorded a mean inference time of $29.7\pm0.7\mathrm{ms}$. While the inference time for MDE can be greatly reduced by running forward passes in parallel \citep{lutjens2018safe}, it will still require significantly more compute resource when compared with PBP, due to the requirement of multiple networks and multiple forward passes. Furthermore, in the case of MDE, the compute time required will increase with the quality requirements of the model uncertainty estimates.

Additionally, the PBP-RNN required far fewer epochs of training to achieve better performance than the MDE, with the PBP-RNN executing a total of 27 epochs of training, while the MDE executed 210 epochs of training.

The PBP-RNN is also advantageous in terms of memory footprint, as it only requires twice the memory of its non-probabilistic alternative, yet provides fully Bayesian uncertainty estimates. In contrast, the quality of the MDE's uncertainty estimates is proportional to the number of networks in the ensemble - requiring larger networks and larger memory footprint to produce uncertainty estimates of reasonable quality \citep{lakshminarayanan2017simple}.

\subsection{Variance Weights for Long Term Memory}
This work demonstrates that a PBP-RNN is capable of achieving superior performance when compared with an LSTM-based approach. This is particularly interesting when considering that traditional RNNs are typically only performant on very short sequences \citep{graves_supervised_2012}, and LSTM's are the state-of-the-art for tasks such as the collision avoidance task used here \citep{lutjens2018safe}. We hypothesize that the performance obtained here is possible due to the variance weights which, through learning variances associated with different features at different time steps, may function similarly to the gates used in an LSTM - however, more rigorous investigation is required to confirm whether this is the case.

\section{Conclusion}
While there has been previous work on Bayesian RNNs \citep{fortunato2017bayesian}, our work is the first to demonstrate that PBP can be effectively applied to an RNN architecture to produce a recurrent network with fully-Bayesian model uncertainty estimates. The resulting network is less demanding on both computation and memory resources than popular dropout and ensemble-based methods, such as the recently proposed MC Dropout Ensemble \citep{lutjens2018safe}. Crucially, our approach produces much higher quality uncertainty estimates, which are necessary for improving RL agent performance in novel scenarios, as demonstrated by the model's competitive performance in dynamic obstacle avoidance tasks. In the case of Safe RL, this directly translates to safer behaviour on the part of the RL agent - a crucial step towards feasibly incorporating more complex machine learning algorithms in safety-critical tasks.


\subsubsection*{Acknowledgements}
Thanks to Clyde Fare, Peter Fenner and Mohab Elkaref for useful discussion. This work was supported by the Science and Technology Facilities Council Hartree Centre’s Innovation Return on Research program, funded by the U.K. Department for Business, Energy, and Industrial Strategy.

\bibliographystyle{apalike}
\bibliography{citations.bib}

\end{document}